\documentclass[10pt,twocolumn]{article}

\usepackage[a4paper,top=19mm,bottom=22mm,left=18mm,right=18mm,columnsep=7mm]{geometry}
\usepackage[T1]{fontenc}
\usepackage[utf8]{inputenc}
\usepackage{lmodern}
\usepackage{amsmath,amssymb,bm}
\usepackage{booktabs}
\usepackage{graphicx}
\usepackage{tikz}
\usetikzlibrary{arrows.meta,calc,positioning}
\usepackage[font=small,labelfont=bf]{caption}
\usepackage{microtype}
\usepackage{xcolor}
\usepackage[hidelinks]{hyperref}

\newcommand{\method}{PromptSpLiCE}

\newcommand{\figref}[1]{Fig.~\ref{#1}}
\newcommand{\tabref}[1]{Table~\ref{#1}}
\newcommand{\eqnref}[1]{Eq.~\eqref{#1}}

\hypersetup{
  pdftitle={Analyzing Learned Prompts in Vision-Language Models through Concept
  Decomposition},
  pdfauthor={Ryo Kamiya, Hiroshi Kera, Kazuhiko Kawamoto}
}

\title{What Does Prompt Learning Change?\\
A Natural-Language Concept Analysis of Vision-Language Models}

\author{
  Ryo Kamiya \quad Hiroshi Kera \quad Kazuhiko Kawamoto\\[0.4em]
  \small Chiba University, Chiba, Japan\\
  \small \texttt{ryo.kamiya@chiba-u.jp},
  \texttt{kera@chiba-u.jp},
  \texttt{kawa@faculty.chiba-u.jp}
}
\date{}

\begin{document}

\maketitle

\begin{abstract}
Prompt learning adapts vision-language models such as CLIP by optimizing continuous prompt
vectors, but the learned prompts are difficult to interpret in natural language. We present
\method{}, a post-hoc method that expresses each class-conditioned text embedding as a
sparse combination of concepts from a fixed natural-language dictionary. Using the same
dictionary before and after prompt learning allows us to compare changes in their concept
profiles. We evaluate \method{} on CoOp, a representative prompt-learning method, across 11
image-classification datasets. The concept profiles change substantially: on average, only
1.6 of the initial top-10 concepts remain in the top 10 after learning. Across datasets,
profile change is positively associated with accuracy gain. 
We also derive a local gradient expression that provides geometric intuition for why
image-aligned concept directions distinct from the current prompt can have greater loss
sensitivity.

\medskip
\noindent\textbf{Keywords:} vision-language models, prompt learning, interpretability,
concept decomposition, CLIP
\end{abstract}

\section{Introduction}
\label{sec:introduction}

Vision-language models map images and text into a shared embedding space, enabling flexible
adaptation through natural-language prompts~\cite{coop,align,blip}. Contrastive
Language-Image Pre-training (CLIP)~\cite{clip}, for example, aligns image and text
representations and supports strong zero-shot transfer to tasks such as image
classification and captioning~\cite{decap}.
The performance of CLIP, however, depends strongly on prompt design. Manual or
large-language-model-based prompt design generally requires costly trial and error. Prompt
learning avoids this process by optimizing a prompt as a sequence of continuous
vectors~\cite{coop,cocoop,kgcoop,lasp}, but makes the resulting prompt difficult to
interpret. For example, a known prompt such as ``A photo of a \{class\}'' may have an
embedding $\bm{z}=(0.30,0.20,\ldots)$, while prompt learning produces
$\bm{z}'=(0.29,0.21,\ldots)$. These coordinate changes do not reveal which concepts have
changed, and $\bm{z}'$ need not correspond to any human-readable sentence.

Existing methods improve interpretability by guiding prompt learning with human-readable
attributes or concepts~\cite{intcoop,xcoop}. They encourage interpretable prompts during
training, but do not provide a general post-hoc interpretation of an arbitrary learned
prompt. Other work decomposes learned representations using sparse autoencoders or
natural-language concepts~\cite{sae,splice}. PatchSAE~\cite{patchsae}, for example, shows
that multimodal prompt learning~\cite{maple} adjusts existing visual features rather than
acquiring entirely new ones. However, many of the resulting features remain difficult to
label in natural language.

To examine these changes in human-readable terms, we represent class-conditioned text
embeddings before and after prompt learning in a shared natural-language coordinate system. 
Using the same concept coordinates before and after learning makes otherwise opaque
embedding changes directly comparable and readable in natural-language terms.
We call this approach \method{}, a prompt-level application of Sparse Linear Concept
Embeddings (SpLiCE)~\cite{splice}. \method{} expresses each prompt embedding as a
nonnegative sparse combination of natural-language concepts, allowing changes induced by
learning to be described through the resulting concept profiles.

We evaluate \method{} on prompts learned by Context Optimization (CoOp)~\cite{coop}, a
representative prompt-learning method, across 11 image-classification datasets. The
analysis reveals substantial reorganization of the natural-language concept profiles: on
average, only 1.6 of the initial top-10 concepts remain in the top 10 after learning.
Qualitative results show that some recognizable class concepts persist even as less
intuitive terms rise in rank. Across datasets, larger profile changes tend to accompany
larger accuracy gains. A local gradient analysis further provides geometric intuition for
why image-aligned concept directions distinct from the current prompt can have greater
local loss sensitivity than directions parallel to it.

Our main contributions are as follows:
\begin{itemize}
  \item We present \method{}, a text-side post-hoc analysis that places initial and learned
  prompt embeddings in a shared coordinate system defined by a fixed natural-language
  dictionary.
  \item We provide a systematic analysis of CoOp across 11 datasets, revealing substantial
  reorganization of fitted concept profiles and illustrating this behavior across all
  datasets.
  \item We derive an embedding-level sensitivity expression that provides geometric
  intuition for loss-sensitive concept directions and distinguish it from the actual CoOp
  parameter gradient.
\end{itemize}

\section{Related Work}
\label{sec:related-work}

\subsection{Prompt Learning}

Context Optimization (CoOp)~\cite{coop} is a representative prompt-learning method for
CLIP. CoOp freezes the pretrained model parameters and learns the token embeddings of a
text prompt. Training maximizes the similarity between an input image and the prompt
associated with its ground-truth class. The learnable token embeddings are initialized with
a manually designed prompt, such as ``A photo of a \{class\}.''

Most prompt-learning methods~\cite{coop,cocoop,kgcoop,lasp,maple} are designed primarily to
improve classification performance and do not explicitly address prompt interpretability.
Recent methods further enrich soft prompts with visual concepts or structured
attributes~\cite{textrefiner,ptincas}, but the resulting task-adapted prompt generally
remains a continuous vector whose semantic change is not directly observable.

Interpretability-aware methods instead guide prompt learning with human-readable knowledge,
such as class attributes or concepts~\cite{intcoop,xcoop,ptincas}. Related work also
constructs interpretable classifiers by aligning local visual regions with language
attributes~\cite{lazsl}. These methods build interpretability into training or inference.
In contrast, we seek to diagnose an already optimized soft prompt post hoc, without
modifying its training objective or the resulting classifier.

\subsection{Concept Decomposition of CLIP Embeddings}

Post-hoc concept decomposition seeks to explain high-dimensional CLIP embeddings in terms
of human-interpretable concepts while keeping the pretrained model fixed. A prominent
approach uses an SAE~\cite{sae} to reconstruct an embedding as a sparse linear combination
of learned basis vectors. Each basis vector can be interpreted from the common properties
of the inputs on which it activates.

This framework has been extended rapidly to vision and vision-language models.
PatchSAE~\cite{patchsae} compares intermediate visual-feature activations before and after
MaPLe adaptation~\cite{maple}, including High$\rightarrow$High, High$\rightarrow$Low, and
Low$\rightarrow$High groups, and tests their influence through top-$k$ latent masking. Our
analysis is complementary: it studies final text embeddings after CoOp learning and uses a
fixed natural-language dictionary, but does not include an intervention test. Later work
evaluated whether SAE features are monosemantic and
human-aligned~\cite{pach2025monosemantic}, aligned concept spaces across different vision
models~\cite{universalsae}, and used SAE features for selective concept
intervention~\cite{sauce} or post-hoc debiasing of CLIP text embeddings~\cite{sem2026}.
Because an SAE learns its dictionary automatically, assigning a precise natural-language
label to every basis vector remains a separate interpretation step.

SpLiCE~\cite{splice} addresses this limitation by using a concept dictionary explicitly
constructed from natural-language terms. It reconstructs CLIP embeddings as sparse
nonnegative combinations of concept embeddings and evaluates the resulting representations
beyond cosine reconstruction. We apply this decomposition procedure to text embeddings
produced by prompt learning and compare the fitted coefficients before and after
optimization. \section{PromptSpLiCE}
\label{sec:method}

Prompt learning adapts a vision-language model by optimizing continuous prompt embeddings,
but changes in these embeddings are difficult to inspect directly. \method{} decomposes the
initial and learned prompt embeddings using a fixed concept dictionary and compares their
fitted coefficients in a common coordinate system.

\begin{figure*}[!t]
  \centering
\begingroup

\newcommand{\PSBars}[3]{%
  \foreach \h [count=\i from 0] in {#2}{
    \pgfmathsetmacro{\xbar}{#1+0.88*\i}
    \pgfmathtruncatemacro{\j}{\i+1}
    \path[bar,fill=#3] (\xbar,0.53) rectangle ++(0.48,\h);
    \node at ({\xbar+0.24},0.31) {$c_{\j}$};
  }
}

\newcommand{\PSActivationProfiles}[3]{%
  \node[font=\bfseries] at (2.12,1.47) {Before};
  \node[font=\bfseries] at (5.05,1.47) {After};

  \begin{scope}[shift={(0,#3)}]
    \draw[gray] (0.72,0.53) -- (3.52,0.53);
    \draw[gray] (3.68,0.53) -- (6.48,0.53);
    \PSBars{1.00}{#1}{blue!70}
    \PSBars{3.96}{#2}{orange!85}
  \end{scope}
}
\newcommand{\PSGeometryBase}{%
  \draw[gray] (8.82,0.95) -- (13.98,0.95);
  \draw[vector,blue] (9.15,0.95) -- (12.48,0.95);
  \node[font=\bfseries,text=blue,anchor=north west]
    at (12.22,0.93) {$\bm z_i$};
  \draw[vector,green!50!black] (9.15,0.95) -- (11.22,1.81);
  \node[font=\bfseries,text=green!50!black,anchor=south east]
    at (11.18,1.70) {$\bm f$};
}

\newcommand{\PSCaseRow}[7]{%
  \begin{scope}[shift={(0,#1)}]
    \draw[panel] (0,#6) rectangle (7.10,2.23);
    \draw[panel] (7.20,#6) rectangle (15.70,2.23);
    \node[tag,anchor=west] at (0.20,1.91) {#2};
    \PSActivationProfiles{#3}{#4}{#7}
    \PSGeometryBase
    #5
  \end{scope}
}
\begin{tikzpicture}[
  x=1cm,y=1cm,
  font=\sffamily\normalsize,
  vector/.style={-{Latex[length=1.8mm,width=1.35mm]},line width=1pt},
  panel/.style={draw=gray,fill=gray!5,rounded corners=1.5mm},
  tag/.style={fill=black,text=white,rounded corners=1.5mm,
    font=\bfseries,inner xsep=1.5mm,inner ysep=0.8mm},
  bar/.style={draw=gray,rounded corners=0.5mm}
]
  \node[font=\bfseries] at (3.45,6.0)
    {CONCEPT ACTIVATIONS};
  \node[font=\bfseries] at (11.65,6.0)
    {GEOMETRIC EXPLANATION};
  \node at (3.45,5.55) {Before vs. after prompt learning};
  \node at (11.65,5.55) {Original embedding coordinates};

  \PSCaseRow{3.05}{LITTLE CHANGE}
    {0.70,0.50,0.28}
    {0.66,0.52,0.30}{%
      \draw[vector,red] (9.15,0.95) -- (12.18,1.13);
      \node[font=\bfseries,text=red,anchor=south west]
        at (11.98,1.08) {$\tilde{\bm c}_a$};
      \node at (11.45,0.32)
        {$\tilde{\bm c}_a\approx\bm z_i
        \;\Longrightarrow\;\bm P_i\tilde{\bm c}_a\approx\bm 0$};
    }{0}{0}

  \PSCaseRow{0.35}{LARGE CHANGE}
    {0.70,0.50,0.28}
    {0.24,0.50,0.76}{%
      \draw[vector,red] (9.15,0.95) -- (11.47,1.67);
      \node[font=\bfseries,text=red,anchor=south west]
        at (11.38,1.5) {$\tilde{\bm c}_b$};
      \node at (11.45,0.05)
        {$\bm P_i\tilde{\bm c}_b\parallel\bm P_i\bm f
        \;\Longrightarrow\;
        \left|\dfrac{\partial\mathcal L}{\partial w_{ib}}\right|
        \ \text{is large}$};
    }{-0.55}{-0.3}
\end{tikzpicture}

\endgroup
  \caption{Illustration of fitted concept-coefficient changes and the
  local geometric intuition considered in this work. A dictionary
  direction parallel to the current prompt has a small tangent
  component, whereas an image-aligned tangent component can have
  greater local loss sensitivity. }
  \label{fig:overview}
\end{figure*}
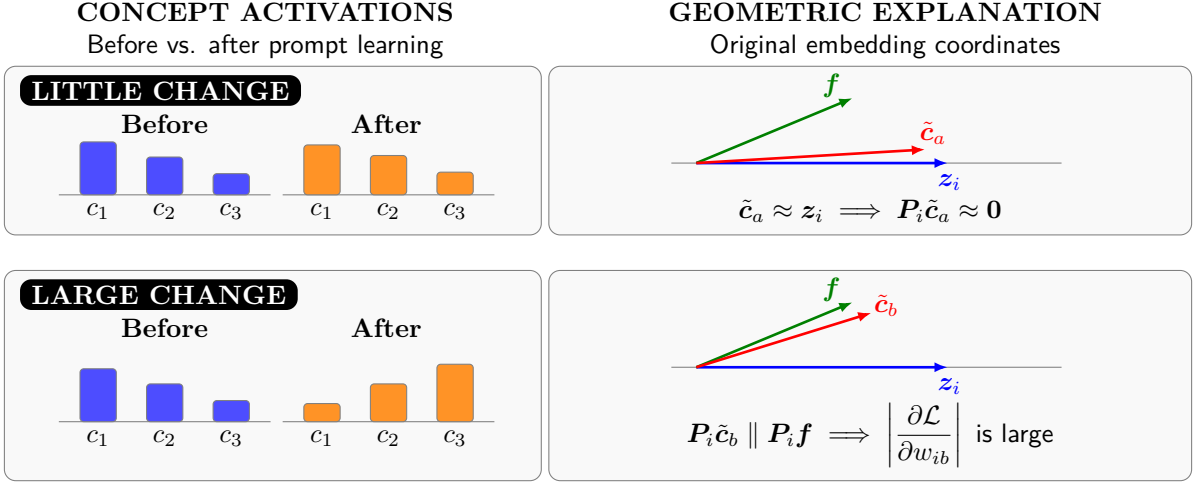

\subsection{Concept Decomposition of Prompt Embeddings}
\label{sec:decomposition}

We define \textbf{\method} as a prompt-level application of Sparse Linear Concept
Embeddings that represents a prompt embedding as a sparse linear combination of a fixed
natural-language dictionary. Decomposing prompts before and after learning in the same
dictionary enables their fitted coefficient profiles to be visualized and compared.

\paragraph{Concept embedding dictionary.}
Let a set of $M$ concepts be specified in natural language. The CLIP text encoder
$g(\cdot)$ maps each concept to an embedding $\bm{c}_j\in\mathbb{R}^{D}$. We arrange these
embeddings as the columns of a concept dictionary
\begin{equation}
  \bm{C}=[\bm{c}_1,\bm{c}_2,\ldots,\bm{c}_M]
  \in\mathbb{R}^{D\times M}.
  \label{eq:concept-dictionary}
\end{equation}

\paragraph{Embedding preprocessing.}
CLIP text embeddings are anisotropic and tend to share a large component in a common
direction~\cite{mind_the_gap}. Under this anisotropy, the common component shared by
concept and prompt embeddings can dominate the reconstruction objective, obscuring
differences among concepts. We mitigate this effect by centering and normalizing the
embeddings. Let
\begin{equation}
  \bm{\mu}=\frac{1}{M}\sum_{j=1}^{M}\bm{c}_j
\end{equation}
be the mean concept embedding. We define
\begin{equation}
  \tilde{\bm{c}}_j=
  \frac{\bm{c}_j-\bm{\mu}}{\lVert\bm{c}_j-\bm{\mu}\rVert_2},
  \quad
  \tilde{\bm{z}}=
  \frac{\bm{z}-\bm{\mu}}{\lVert\bm{z}-\bm{\mu}\rVert_2},
  \label{eq:preprocessing}
\end{equation}
and replace \eqnref{eq:concept-dictionary} by
\begin{equation}
  \tilde{\bm{C}}=
  [\tilde{\bm{c}}_1,\tilde{\bm{c}}_2,\ldots,\tilde{\bm{c}}_M].
\end{equation}

\paragraph{Concept-coefficient estimation.}
We approximate $\tilde{\bm{z}}$ by a linear combination of the centered concept embeddings.
Let $\bm{w}\in\mathbb{R}_{\geq 0}^{M}$ be a nonnegative vector of decomposition
coefficients. We solve
\begin{equation}
  \begin{aligned}
  \bm{w}^{*}=\operatorname*{arg\,min}_{\bm{w}\geq 0}\;
  \frac{1}{2}\lVert\tilde{\bm{C}}\bm{w}-\tilde{\bm{z}}\rVert_2^2
  +\lambda\lVert\bm{w}\rVert_1,
  \end{aligned}
  \label{eq:nonnegative-lasso}
\end{equation}
where $\lambda$ controls the trade-off between reconstruction error and sparsity. The
$\ell_1$ penalty sets many coefficients to zero. The coefficients are fitted post hoc and
are not activations measured inside CLIP. Because the dictionary is overcomplete and
contains correlated word embeddings, similar reconstructions may admit different
coefficient supports. Equation~\eqref{eq:preprocessing} removes the length of the centered
residual. In particular, if
\begin{equation}
  \rho_{\bm{z}}=\lVert\bm{z}-\bm{\mu}\rVert_2,
  \quad
  \bm{z}=\bm{\mu}+\rho_{\bm{z}}\tilde{\bm{z}},
  \label{eq:inverse-preprocessing}
\end{equation}
then an exact inversion requires both the direction $\tilde{\bm{z}}$ and the scale
$\rho_{\bm{z}}$. The formulation used here retains the direction and fixes the residual
scale to one. Assuming that $\bm{z}$ has unit norm, the resulting direction-only
reconstruction is
\begin{equation}
  \bm{z}\approx
  \frac{\tilde{\bm{C}}\bm{w}^{*}+\bm{\mu}}
  {\lVert\tilde{\bm{C}}\bm{w}^{*}+\bm{\mu}\rVert_2}.
  \label{eq:reconstruction}
\end{equation}
Appendix~\ref{app:reconstruction-derivation} gives the corresponding scale-aware expression
and makes explicit the approximation introduced by this convention.

\subsection{Local Sensitivity in Concept Coordinates}
\label{sec:theory}

We next analyze a hypothetical local sensitivity of a cosine-similarity-based
classification objective in concept coordinates. We regard $\bm{w}_i$ as a local coordinate
vector for the reconstructed text embedding of class $i$, while holding the image embedding
and the other class embeddings fixed. In this setting, prompt learning does not optimize
$\bm{w}_i$ directly; it optimizes prompt parameters, and \method{} fits $\bm{w}_i$
afterward. Thus, the following derivative is an embedding-level sensitivity rather than the
gradient with respect to the prompt parameters or the total derivative of the fitted Lasso
solution. For the prompt associated with class $i$, the chain rule gives
\begin{equation}
  \frac{\partial\mathcal{L}}{\partial w_{ij}}
  =
  \frac{\partial\mathcal{L}}{\partial s_i}
  \frac{\partial s_i}{\partial w_{ij}},
  \label{eq:chain-rule}
\end{equation}
where $\mathcal{L}$ is the cross-entropy loss, $w_{ij}$ is the fitted coefficient of
dictionary term $j$ for class $i$, and $s_i$ is the cosine similarity for class $i$.

Under the direction-only reconstruction convention in \eqnref{eq:reconstruction}, let
$\bm{f}\in\mathbb{R}^{D}$ be a unit-normalized image embedding and define
\begin{equation}
  \bm{u}_i=\tilde{\bm{C}}\bm{w}_i+\bm{\mu},
  \quad
  \bm{z}_i=\frac{\bm{u}_i}{\lVert\bm{u}_i\rVert_2},
  \quad
  s_i=\bm{f}^{\top}\bm{z}_i.
  \label{eq:similarity}
\end{equation}
The class probabilities are
\begin{equation}
  p_i=
  \frac{\exp(s_i/\tau)}{\sum_k\exp(s_k/\tau)},
\end{equation}
where $\tau$ is the softmax temperature. For a one-hot target $\bm{t}$ and cross-entropy
loss $\mathcal{L}=-\sum_k t_k\log p_k$, the softmax derivative is
\begin{equation}
  \frac{\partial\mathcal{L}}{\partial s_i}
  =\frac{p_i-t_i}{\tau}.
  \label{eq:score-gradient}
\end{equation}
The remaining derivatives are
\begin{equation}
  \begin{aligned}
  \frac{\partial s_i}{\partial\bm{z}_i}
  =\bm{f}^{\top},
  \frac{\partial\bm{z}_i}{\partial\bm{u}_i}
  =\frac{1}{\lVert\bm{u}_i\rVert_2}
    \left(\bm{I}-\bm{z}_i\bm{z}_i^{\top}\right),
  \frac{\partial\bm{u}_i}{\partial w_{ij}}
  =\tilde{\bm{c}}_j.
  \end{aligned}
   \label{eq:normalization-jacobian}
\end{equation}
Combining \eqnref{eq:chain-rule}, \eqnref{eq:score-gradient}, and
\eqnref{eq:normalization-jacobian} gives
\begin{equation}
  \frac{\partial\mathcal{L}}{\partial w_{ij}}
  =
  \frac{p_i-t_i}{\tau\lVert\bm{u}_i\rVert_2}
  \bm{f}^{\top}
  \left(\bm{I}-\bm{z}_i\bm{z}_i^{\top}\right)
  \tilde{\bm{c}}_j.
  \label{eq:activation-gradient}
\end{equation}
The complete differential calculation and its relation to the underlying prompt parameters
are provided in Appendix~\ref{app:gradient-derivation}.

The factor $p_i-t_i$ in Eq.~(\ref{eq:normalization-jacobian}) reflects the prediction
error. For the ground-truth class, it is $p_i-1$, whose magnitude is larger when the
correct-class probability is low and approaches zero as the prediction becomes confident.
The temperature $\tau$ and the norm $\lVert\bm{u}_i\rVert_2$ scale the update without
changing its qualitative dependence on the concept direction.
The direction-dependent term in \eqnref{eq:activation-gradient} can be written as
\begin{equation}
  \bm{f}^{\top}
  \left(\bm{I}-\bm{z}_i\bm{z}_i^{\top}\right)
  \tilde{\bm{c}}_j
  =
  \bm{f}^{\top}\tilde{\bm{c}}_j
  -s_i\bm{z}_i^{\top}\tilde{\bm{c}}_j.
  \label{eq:projection-term}
\end{equation}
The matrix $\bm{I}-\bm{z}_i\bm{z}_i^{\top}$ projects onto the subspace orthogonal to the
current prompt embedding. Equation~\eqref{eq:projection-term} therefore measures how
strongly the component of dictionary direction $j$ orthogonal to the prompt aligns with the
image embedding.
If $\tilde{\bm{c}}_j$ is nearly parallel to $\bm{z}_i$, its projected component and the
hypothetical derivative with respect to $w_{ij}$ are small.

Figure~\ref{fig:overview} illustrates when concept changes can be small or large in this
local view. For comparable prediction-error and scale factors, a concept changes little
when its dictionary direction is nearly parallel to the current prompt embedding
$\bm{z}_i$, because its component orthogonal to $\bm{z}_i$ is small (upper). In contrast, a
concept can change substantially when its orthogonal component aligns with that of the
image embedding $\bm{f}$, resulting in high local loss sensitivity (lower).

\section{Experiments}
\label{sec:experiments}

We evaluate \method{} as a post-hoc natural-language decomposition of prompt embeddings
before and after learning. We first assess reconstruction of the full coefficient vector
and then examine changes in its ranking and distribution.

\subsection{Experimental Setup}
\label{sec:setup}

\paragraph{Datasets.}
We use 11 image-classification datasets: ImageNet~\cite{imagenet},
OxfordPets~\cite{oxfordpets}, Caltech101~\cite{caltech101},
StanfordCars~\cite{stanfordcars}, Food101~\cite{food101}, Flowers102~\cite{flowers102},
FGVCAircraft~\cite{fgvcaircraft}, SUN397~\cite{sun397}, DTD~\cite{dtd},
EuroSAT~\cite{eurosat}, and UCF101~\cite{ucf101}. Together, they cover generic object
recognition, fine-grained classification, scene recognition, action recognition, texture
recognition, and satellite-image classification.

\begin{figure}[tb]
  \centering
  \includegraphics[width=\columnwidth]{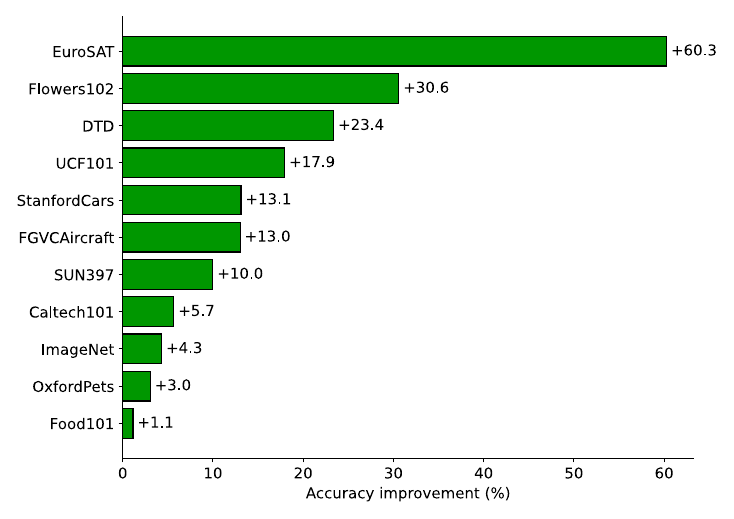}
  \caption{Improvement in classification accuracy obtained by CoOp relative to the initial
  CLIP prompt.}
  \label{fig:accuracy-improvement}
\end{figure}

\paragraph{Vision-language model and prompt learning.}
We use CLIP~\cite{clip} with a ResNet-50 image encoder~\cite{resnet50}; both the image and
text encoders are pretrained and frozen. We apply CoOp~\cite{coop}, initialize its context
with ``A photo of a \{class\},'' and train it with 16 images per class. The context tokens
are optimized by SGD for 200 epochs with a batch size of 32 and a base learning rate of
$2\times10^{-3}$. The first epoch uses a warm-up learning rate of $1\times10^{-5}$,
followed by cosine annealing. As shown in \figref{fig:accuracy-improvement}, this setting
improves classification accuracy on every dataset. The largest gain is 60.3 percentage
points on EuroSAT. All reported analyses use this single RN50--CoOp configuration.

\paragraph{Concept decomposition.}
Following SpLiCE~\cite{splice}, we construct the concept dictionary from the 10,000 most
frequent words extracted from LAION-400M captions~\cite{Schuhmann2021LAION400MOD}. The
frequency-derived vocabulary contains misspellings and other noisy lexical items, which
limits the readability of some fitted labels. The mean of the concept embeddings is used to
center both concept and prompt embeddings. We solve \eqnref{eq:nonnegative-lasso} using the
alternating direction method of multipliers (ADMM)~\cite{admm} and set the $\ell_1$
regularization coefficient to $\lambda=0.01$ unless otherwise stated.

\subsection{Embedding Reconstruction Fidelity}
\label{sec:reconstruction}

We first evaluate how closely the full \method{} decomposition reconstructs an original
prompt embedding $\bm{z}$. The evaluation metric is the cosine similarity between $\bm{z}$
and its reconstruction from \eqnref{eq:reconstruction}:
\begin{equation}
  \cos\!\left(
    \bm{z},
    \frac{\tilde{\bm{C}}\bm{w}^{*}+\bm{\mu}}
    {\lVert\tilde{\bm{C}}\bm{w}^{*}+\bm{\mu}\rVert_2}
  \right).
  \label{eq:reconstruction-metric}
\end{equation}
A value close to one indicates directional agreement between the original embedding and the
reconstruction from the full coefficient vector. It does not establish that the top-ranked
terms alone faithfully represent the prompt or preserve its predictions.

\begin{figure}[tb]
  \centering
  \includegraphics[width=\columnwidth]{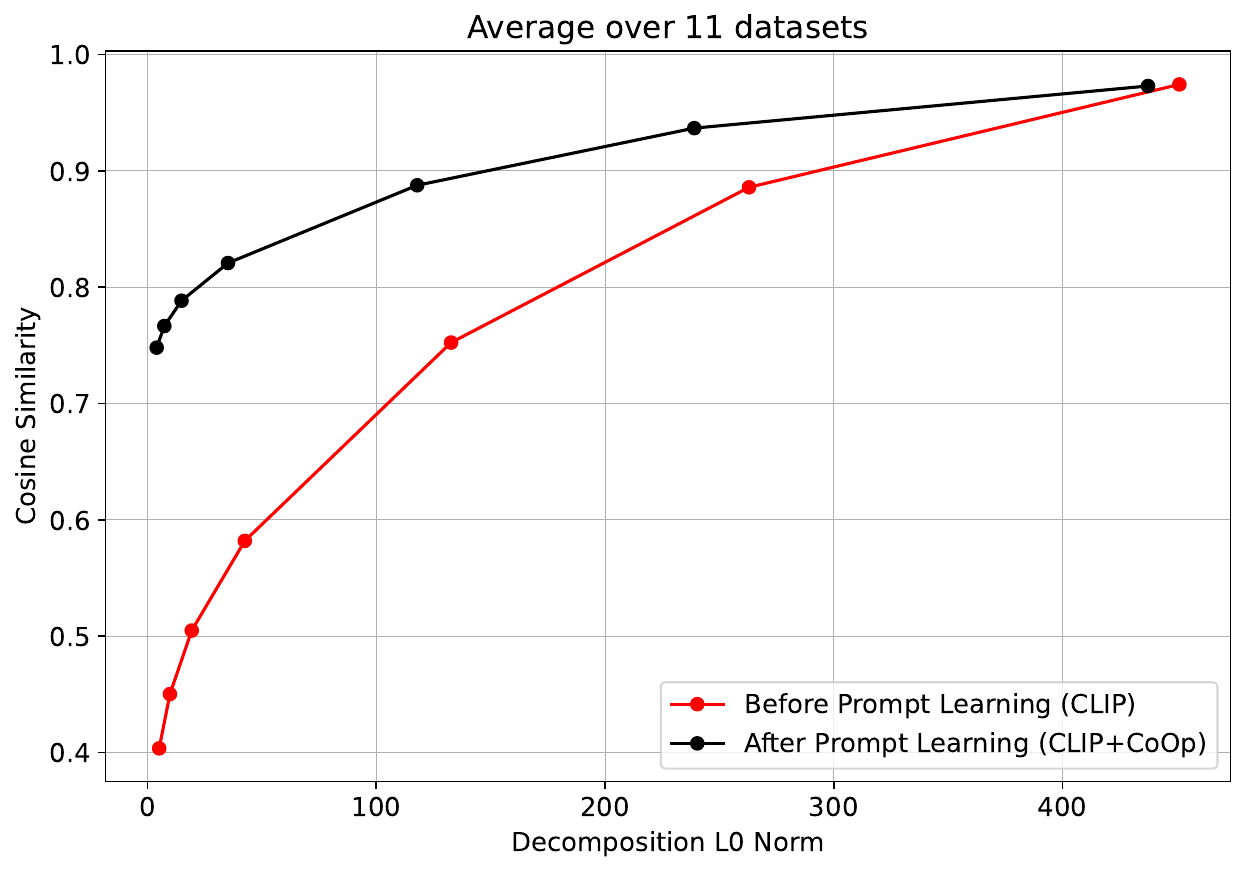}
  \caption{Cosine similarity between original prompt embeddings and their full \method{}
  reconstructions, averaged over 11 datasets. Sparsity is controlled by
  $\lambda\in\{0.5,0.4,0.3,0.2,0.1,0.05,0.01\}$. }
  \label{fig:reconstruction-fidelity}
\end{figure}

Figure~\ref{fig:reconstruction-fidelity} reports the average over the 11 datasets for
prompts before and after learning. We vary the regularization coefficient in
\eqnref{eq:nonnegative-lasso} over $\lambda\in\{0.5,0.4,0.3,0.2,0.1,0.05,0.01\}$. The
horizontal axis shows the number of nonzero coefficients for each value of $\lambda$. At
$\lambda=0.01$, the learned prompts have approximately 450 nonzero coefficients and achieve
a cosine similarity of 0.98. Although this is sparse relative to the 10,000-term
dictionary, 450 terms do not constitute a concise explanation. The top-10 displays below
are illustrative subsets of this larger solution; their cumulative mass and reconstruction
fidelity are not evaluated here.

\begin{figure}[tb]
  \centering
  \includegraphics[width=\columnwidth]{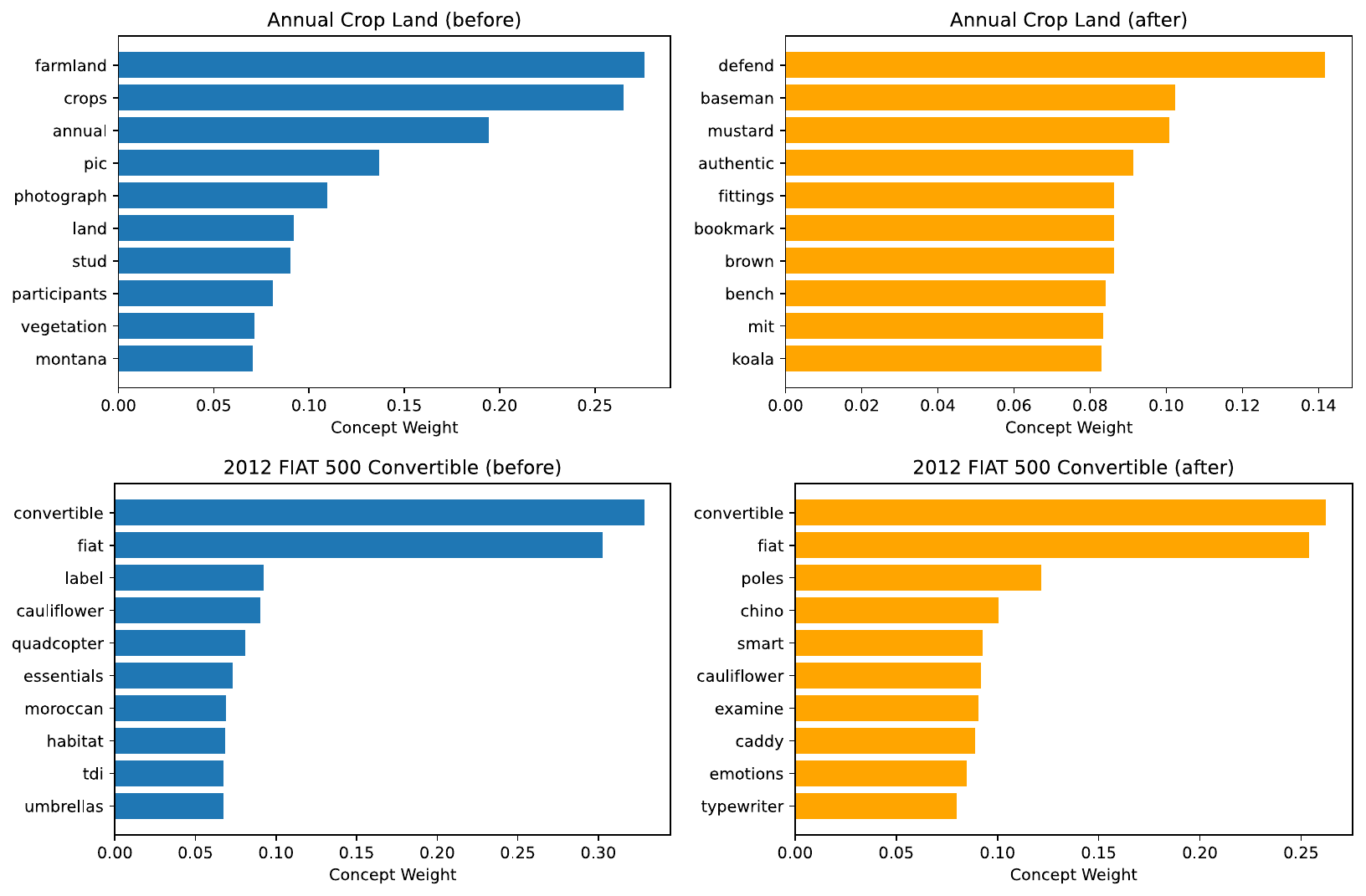}
  \caption{Top-10 dictionary terms and their fitted coefficients before and after prompt
  learning. The top row shows EuroSAT's ``Annual Crop Land'' class, and the bottom row
  shows StanfordCars' ``2012 FIAT 500 Convertible'' class.}
  \label{fig:concept-bars}
\end{figure}

Figure~\ref{fig:concept-bars} shows example decompositions for the EuroSAT class ``Annual
Crop Land'' and the StanfordCars class ``2012 FIAT 500 Convertible.'' For Annual Crop Land,
\emph{farmland} and \emph{crops} have high fitted coefficients for the initial prompt,
whereas \emph{defend} and \emph{baseman} rank highly for the learned prompt. 

\subsection{Changes in Coefficient Profiles}
\label{sec:concept-changes}

We compare the fitted coefficient rankings before and after prompt learning. These rankings
are obtained with a fixed dictionary, solver, and regularization value and should not be
interpreted as direct measurements of internal concept activation.

\subsubsection{Quantitative Analysis}

Each dictionary term is assigned to one of three groups:
\begin{itemize}
  \setlength{\itemsep}{0.2em}
  \item \textbf{High$\rightarrow$High:} ranked in the top 10 both before and after
  learning;
  \item \textbf{High$\rightarrow$Low:} ranked in the top 10 before learning but below the
  top 100 after learning; and
  \item \textbf{Low$\rightarrow$High:} ranked below the top 100 before learning but in the
  top 10 after learning.
\end{itemize}
The top-10 and below-100 cutoffs provide a simple descriptive summary, but they are
threshold choices; crossing a boundary does not by itself imply that a model-internal
concept was suppressed or newly acquired.

Table~\ref{tab:rank-transitions} reports the average group size across classes for each
dataset. High$\rightarrow$High terms are few on most datasets, while High$\rightarrow$Low
and Low$\rightarrow$High transitions are common. Averaged over all 11 datasets, only 1.6 of
the initial top-10 terms remain in the top 10; 6.8 fall below rank 100, while 6.8 rise from
below rank 100 into the top 10. 

EuroSAT exhibits the largest change in the fitted coefficient rankings, with averages of
0.3, 9.2, and 9.4 for High$\rightarrow$High, High$\rightarrow$Low, and
Low$\rightarrow$High, respectively. 
Limited coverage of satellite-domain terms in the general-purpose LAION-derived dictionary
may contribute to this result. Across datasets, the fitted rankings change substantially.
However, because correlated dictionary terms can replace one another in the Lasso solution,
these changes may arise from both CoOp-induced changes and instability in the
decomposition.

\begin{table}[tb]
  \centering
  \caption{Average number of dictionary terms in each rank-transition group.}
  \label{tab:rank-transitions}
  \resizebox{\columnwidth}{!}{%
  \begin{tabular}{lccc}
    \toprule
    Dataset & High$\rightarrow$High & High$\rightarrow$Low & Low$\rightarrow$High \\
    \midrule
    EuroSAT      & 0.3 & 9.2 & 9.4 \\
    DTD          & 0.8 & 8.3 & 8.5 \\
    FGVCAircraft & 1.1 & 7.3 & 7.2 \\
    UCF101       & 1.2 & 7.8 & 7.8 \\
    Caltech101   & 1.5 & 7.1 & 7.2 \\
    Food101      & 1.6 & 6.3 & 6.5 \\
    Flowers102   & 1.9 & 6.0 & 5.9 \\
    ImageNet     & 1.9 & 6.2 & 6.9 \\
    SUN397       & 1.9 & 6.6 & 6.6 \\
    StanfordCars & 2.5 & 4.8 & 4.7 \\
    OxfordPets   & 3.1 & 4.7 & 4.5 \\
    \midrule
    Average      & 1.6 & 6.8 & 6.8 \\
    \bottomrule
  \end{tabular}%
  }
\end{table}

\begin{figure}[tb]
  \centering
  \includegraphics[
    width=\columnwidth,
    trim=0 270bp 0 0,
    clip
  ]{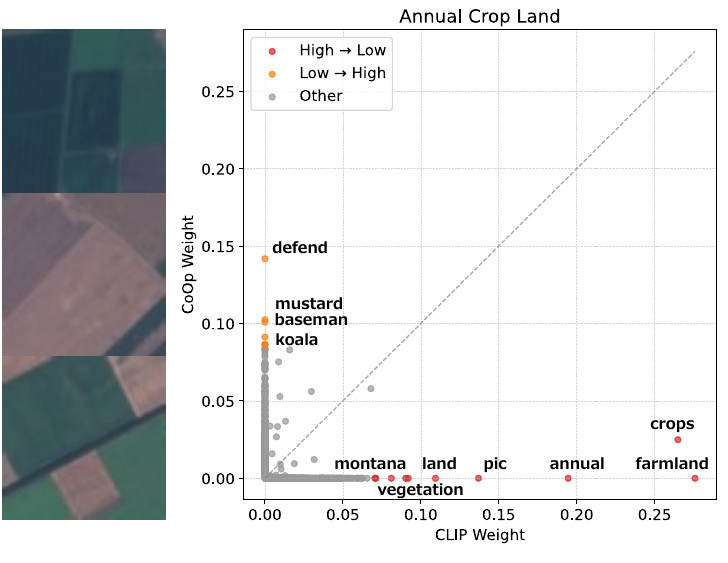}
  {\small (a) EuroSAT: ``Annual Crop Land''\par}

  \vspace{0.6em}
  \includegraphics[
    width=\columnwidth,
    trim=0 0 0 270bp,
    clip
  ]{figures/casestudy.pdf}
  {\small (b) StanfordCars: ``2012 FIAT 500 Convertible''\par}

  \caption{Changes in fitted decomposition coefficients for representative classes. Each
  point is a dictionary term; the horizontal and vertical axes show its coefficient before
  (CLIP) and after (CoOp) prompt learning, respectively. Selected High$\rightarrow$High,
  High$\rightarrow$Low, and Low$\rightarrow$High terms are labeled.}
  \label{fig:case-study}
\end{figure}

\subsubsection{Qualitative Analysis}

Figure~\ref{fig:case-study} compares fitted coefficient rankings before and after learning,
showing larger changes for EuroSAT than for StanfordCars.
For EuroSAT's ``Annual Crop Land'' class, directly related terms such as \emph{farmland}
and \emph{crops} move from the top 10 to below rank 100, whereas \emph{defend} and
\emph{baseman} make the opposite transition. For StanfordCars' ``2012 FIAT 500
Convertible'' class, class-related terms such as \emph{fiat} and \emph{convertible} remain
highly ranked, while less intuitive terms such as \emph{poles} and \emph{typewriter} move
upward.

\begin{figure}[tb]
  \centering
  \includegraphics[width=\columnwidth]{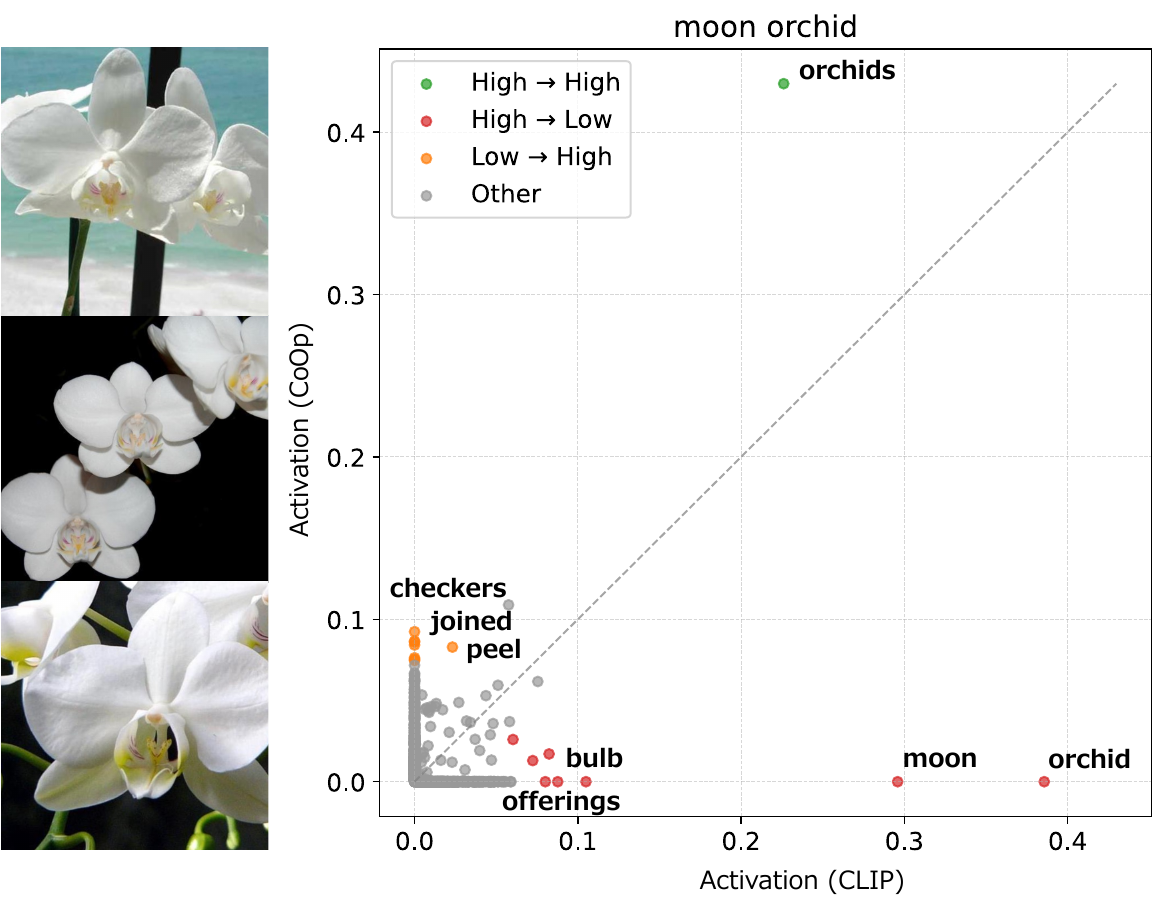}
  {\small (a) Flowers102: ``moon orchid''\par}
  \vspace{0.6em}
  \includegraphics[width=\columnwidth]{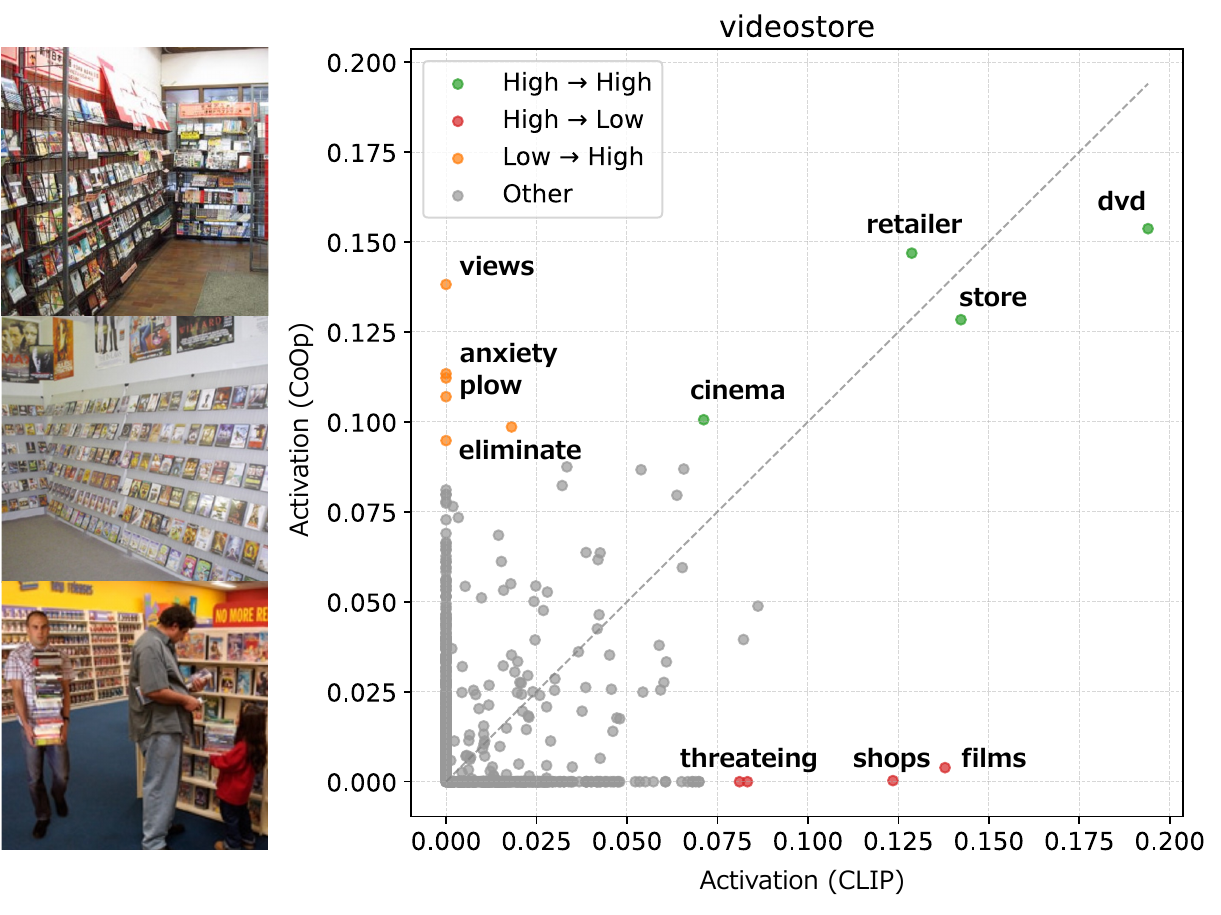}
  {\small (b) SUN397: ``videostore''\par}
  \caption{Complementary decomposition-coefficient changes for Flowers102 and SUN397. The
  axes and colors follow \figref{fig:case-study}.}
  \label{fig:main-additional-cases}
\end{figure}

Figure~\ref{fig:main-additional-cases} shows two complementary cases. For Flowers102's
``moon orchid'' class, \emph{orchids} remains highly ranked, but \emph{moon},
\emph{orchid}, and \emph{bulb} move downward; \emph{checkers}, \emph{joined}, and
\emph{peel} move upward instead. For SUN397's ``videostore'' class, a coherent set of
scene-related terms, including \emph{dvd}, \emph{store}, \emph{retailer}, and
\emph{cinema}, retains high coefficients, while \emph{shops} and \emph{films} move downward
and less intuitive terms move upward.

Taken together, these examples show that the selected dictionary profile does not
necessarily preserve the rank of terms that humans regard as semantically related to a
class, nor does it uniformly move them downward. Appendix~\ref{app:additional-cases}
provides corresponding examples for the remaining seven datasets.

\subsection{Exploratory Association with Accuracy Gain}
\label{sec:distribution-and-accuracy}

We next quantify the magnitude of the coefficient-distribution change and compare it with
the accuracy gain from prompt learning. Let $P$ and $Q$ be the normalized nonnegative
\method{} coefficient distributions before and after learning, respectively. We measure
their difference with the Jensen--Shannon (JS) divergence:
\begin{equation}
  \operatorname{JS}(P\parallel Q)
  =\frac{1}{2}\operatorname{KL}(P\parallel M)
  +\frac{1}{2}\operatorname{KL}(Q\parallel M),
  \label{eq:js-divergence}
\end{equation}
where $M=\tfrac{1}{2}(P+Q)$ and $\operatorname{KL}$ denotes the Kullback--Leibler
divergence.

\begin{figure}[tb]
  \centering
  \includegraphics[width=\columnwidth]{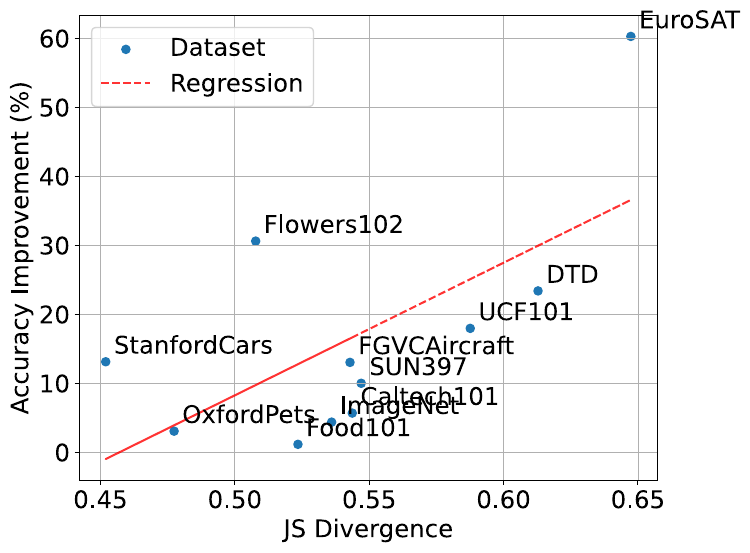}
  \caption{Mean fitted coefficient-distribution change versus classification-accuracy
  improvement across the 11 datasets (Pearson's $r=0.64$). The dashed line is a linear
  regression fit. The 11-point association is exploratory.}
  \label{fig:js-and-accuracy}
\end{figure}

Figure~\ref{fig:js-and-accuracy} plots, for each of the 11 datasets, the mean JS divergence
between initial and learned prompts against the corresponding accuracy gain from CoOp. The
dataset-level Pearson correlation is $r=0.64$ (nominal $p=0.035$, $n=11$). 
We treat this as an exploratory association in the present configuration. The small sample
and the prominent EuroSAT endpoint make the estimate potentially sensitive to individual
datasets.

\section{Conclusion}
\label{sec:conclusion}

We present \method{}, a post-hoc method that places class-conditioned text embeddings
before and after prompt learning in a shared coordinate system defined by a fixed
natural-language dictionary. Applied to CoOp across 11 datasets, the analysis revealed
substantial changes in the fitted concept profiles. Larger profile changes also tended to
accompany larger accuracy gains. A local gradient analysis provided geometric intuition for
why image-aligned concept directions distinct from the current prompt can have greater loss
sensitivity.

\method{} makes changes in continuous prompts readable through named concepts without
modifying the prompt-learning procedure. It therefore provides a practical text-side
diagnostic that can be applied beyond CoOp and complements analyses of visual
representations.

\section*{Acknowledgments}
This work was supported by JSPS KAKENHI Grant Number 26K23832.

\bibliographystyle{unsrt}
\bibliography{references}

@inproceedings{clip,
  author       = {Alec Radford and
                  others},
  title        = {Learning Transferable Visual Models From Natural Language Supervision},
  booktitle    = {International Conference on Machine Learning},
  pages        = {8748--8763},
  year         = {2021},
}

@inproceedings{align,
  author       = {Chao Jia and others},
  title        = {Scaling Up Visual and Vision-Language Representation Learning With Noisy Text Supervision},
  booktitle    = {International Conference on Machine Learning},
  pages        = {4904--4916},
  year         = {2021},
}

@inproceedings{blip,
  author       = {Junnan Li and
                  Dongxu Li and
                  Caiming Xiong and
                  Steven C. H. Hoi},
  title        = {{BLIP}: Bootstrapping Language-Image Pre-training for Unified Vision-Language Understanding and Generation},
  booktitle    = {International Conference on Machine Learning},
  pages        = {12888--12900},
  year         = {2022},
}

@article{coop,
  author       = {Kaiyang Zhou and
                  Jingkang Yang and
                  Chen Change Loy and
                  Ziwei Liu},
  title        = {Learning to Prompt for Vision-Language Models},
  journal      = {International Journal of Computer Vision},
  volume       = {130},
  number       = {9},
  pages        = {2337--2348},
  year         = {2022},
}

@inproceedings{cocoop,
  author       = {Kaiyang Zhou and
                  Jingkang Yang and
                  Chen Change Loy and
                  Ziwei Liu},
  title        = {Conditional Prompt Learning for Vision-Language Models},
  booktitle    = {Computer Vision and Pattern Recognition},
  pages        = {16816--16825},
  year         = {2022},
}

@inproceedings{maple,
  author       = {Muhammad Uzair Khattak and
                  Hanoona Abdul Rasheed and
                  Muhammad Maaz and
                  Salman H. Khan and
                  Fahad Shahbaz Khan},
  title        = {{MaPLe}: Multi-modal Prompt Learning},
  booktitle    = {Computer Vision and Pattern Recognition},
  pages        = {19113--19122},
  year         = {2023},
}

@inproceedings{kgcoop,
  author       = {Hantao Yao and
                  Rui Zhang and
                  Changsheng Xu},
  title        = {Visual-Language Prompt Tuning with Knowledge-Guided Context Optimization},
  booktitle    = {Computer Vision and Pattern Recognition},
  pages        = {6757--6767},
  year         = {2023},
}

@inproceedings{lasp,
  author       = {Adrian Bulat and
                  Georgios Tzimiropoulos},
  title        = {{LASP}: Text-to-Text Optimization for Language-Aware Soft Prompting of Vision \& Language Models},
  booktitle    = {Computer Vision and Pattern Recognition},
  pages        = {23232--23241},
  year         = {2023},
}

@inproceedings{intcoop,
  author       = {Soumya Suvra Ghosal and
                  Samyadeep Basu and
                  Soheil Feizi and
                  Dinesh Manocha},
  title        = {{IntCoOp}: Interpretability-Aware Vision-Language Prompt Tuning},
  booktitle    = {Empirical Methods in Natural Language Processing},
  pages        = {19584--19601},
  year         = {2024},
}

@inproceedings{xcoop,
  author       = {Yequan Bie and
                  Luyang Luo and
                  Zhixuan Chen and
                  Hao Chen},
  title        = {{XCoOp}: Explainable Prompt Learning for Computer-Aided Diagnosis via Concept-Guided Context Optimization},
  booktitle    = {Medical Image Computing and Computer Assisted Intervention},
  pages        = {773--783},
  year         = {2024},
}

@inproceedings{splice,
  author       = {Usha Bhalla and
                  Alex Oesterling and
                  Suraj Srinivas and
                  Fl{\'{a}}vio P. Calmon and
                  Himabindu Lakkaraju},
  title        = {Interpreting {CLIP} with Sparse Linear Concept Embeddings ({SpLiCE})},
  booktitle    = {Neural Information Processing Systems},
  volume       = {37},
  year         = {2024},
  doi          = {10.52202/079017-2678},
}

@inproceedings{sae,
  author       = {Robert Huben and
                  Hoagy Cunningham and
                  Logan Riggs Smith and
                  Aidan Ewart and
                  Lee Sharkey},
  title        = {Sparse Autoencoders Find Highly Interpretable Features in Language Models},
  booktitle    = {International Conference on Learning Representations},
  year         = {2024},
}

@inproceedings{patchsae,
  author       = {Hyesu Lim and
                  Jinho Choi and
                  Jaegul Choo and
                  Steffen Schneider},
  title        = {Sparse autoencoders reveal selective remapping of visual concepts during adaptation},
  booktitle    = {International Conference on Learning Representations},
  year         = {2025},
}

@inproceedings{textrefiner,
  author       = {Jingjing Xie and
                  Yuxin Zhang and
                  Jun Peng and
                  Zhaohong Huang and
                  Liujuan Cao},
  title        = {{TextRefiner}: Internal Visual Feature as Efficient Refiner for Vision-Language Models Prompt Tuning},
  booktitle    = {Proceedings of the AAAI Conference on Artificial Intelligence},
  volume       = {39},
  pages        = {8718--8726},
  year         = {2025},
  doi          = {10.1609/aaai.v39i8.32942},
}

@inproceedings{ptincas,
  author       = {Shiyu Hou and
                  Tianfei Zhou and
                  Shuai Zhang and
                  Ye Yuan and
                  Guoren Wang},
  title        = {Prompt Tuning in a Compact Attribute Space},
  booktitle    = {Proceedings of the AAAI Conference on Artificial Intelligence},
  volume       = {39},
  pages        = {3518--3526},
  year         = {2025},
  doi          = {10.1609/aaai.v39i4.32365},
}

@inproceedings{lazsl,
  author       = {Shiming Chen and
                  Bowen Duan and
                  Salman Khan and
                  Fahad Shahbaz Khan},
  title        = {Interpretable Zero-Shot Learning with Locally-Aligned Vision-Language Model},
  booktitle    = {International Conference on Computer Vision},
  pages        = {478--487},
  year         = {2025},
}

@inproceedings{pach2025monosemantic,
  author       = {Mateusz Pach and
                  Shyamgopal Karthik and
                  Quentin Bouniot and
                  Serge Belongie and
                  Zeynep Akata},
  title        = {Sparse Autoencoders Learn Monosemantic Features in Vision-Language Models},
  booktitle    = {Neural Information Processing Systems},
  volume       = {38},
  year         = {2025},
  doi          = {10.52202/085713-3202},
}

@inproceedings{universalsae,
  author       = {Harrish Thasarathan and
                  Julian Forsyth and
                  Thomas Fel and
                  Matthew Kowal and
                  Konstantinos G. Derpanis},
  title        = {Universal Sparse Autoencoders: Interpretable Cross-Model Concept Alignment},
  booktitle    = {International Conference on Machine Learning},
  volume       = {267},
  pages        = {59304--59325},
  year         = {2025},
}

@inproceedings{sauce,
  author       = {Jiahui Geng and
                  Qing Li},
  title        = {{SAUCE}: Selective Concept Unlearning in Vision-Language Models with Sparse Autoencoders},
  booktitle    = {International Conference on Computer Vision},
  pages        = {3023--3033},
  year         = {2025},
}

@inproceedings{sem2026,
  author       = {Quentin Guimard and
                  Federico Bartsch and
                  Simone Caldarella and
                  Rahaf Aljundi and
                  Elisa Ricci and
                  Massimiliano Mancini},
  title        = {{SEM}: Sparse Embedding Modulation for Post-Hoc Debiasing of Vision-Language Models},
  booktitle    = {Computer Vision and Pattern Recognition Findings},
  pages        = {8101--8110},
  year         = {2026},
}

@inproceedings{mind_the_gap,
  author       = {Victor Weixin Liang and
                  Yuhui Zhang and
                  Yongchan Kwon and
                  Serena Yeung and
                  James Y. Zou},
  title        = {Mind the Gap: Understanding the Modality Gap in Multi-modal Contrastive Representation Learning},
 booktitle    = {Neural Information Processing Systems},
  volume       = {35},
  pages        = {17612--17625},
  year         = {2022},
}

@article{admm,
  author       = {Stephen P. Boyd and
                  Neal Parikh and
                  Eric Chu and
                  Borja Peleato and
                  Jonathan Eckstein},
  title        = {Distributed Optimization and Statistical Learning via the Alternating Direction Method of Multipliers},
  journal      = {Foundations and Trends in Machine Learning},
  volume       = {3},
  number       = {1},
  pages        = {1--122},
  year         = {2011},
}

@inproceedings{decap,
  author       = {Wei Li and
                  Linchao Zhu and
                  Longyin Wen and
                  Yi Yang},
  title        = {{DeCap}: Decoding {CLIP} Latents for Zero-Shot Captioning via Text-Only Training},
  booktitle    = {International Conference on Learning Representations},
  year         = {2023},
}

@inproceedings{resnet50,
  author       = {Kaiming He and
                  Xiangyu Zhang and
                  Shaoqing Ren and
                  Jian Sun},
  title        = {Deep Residual Learning for Image Recognition},
  booktitle    = {Computer Vision and Pattern Recognition},
  pages        = {770--778},
  year         = {2016},
}

@inproceedings{imagenet,
  author       = {Jia Deng and
                  Wei Dong and
                  Richard Socher and
                  Li{-}Jia Li and
                  Kai Li and
                  Li Fei{-}Fei},
  title        = {{ImageNet}: A large-scale hierarchical image database},
  booktitle    = {Computer Vision and Pattern Recognition},
  pages        = {248--255},
  year         = {2009},
}

@inproceedings{caltech101,
  author       = {Li Fei{-}Fei and
                  Rob Fergus and
                  Pietro Perona},
  title        = {Learning Generative Visual Models from Few Training Examples: An Incremental Bayesian Approach Tested on 101 Object Categories},
  booktitle    = {Computer Vision and Pattern Recognition Workshops},
  pages        = {178},
  year         = {2004},
}

@article{eurosat,
  author       = {Patrick Helber and
                  Benjamin Bischke and
                  Andreas Dengel and
                  Damian Borth},
  title        = {{EuroSAT}: A Novel Dataset and Deep Learning Benchmark for Land Use and Land Cover Classification},
  journal      = {IEEE Journal of Selected Topics in Applied Earth Observations and Remote Sensing},
  volume       = {12},
  number       = {7},
  pages        = {2217--2226},
  year         = {2019},
}

@article{ucf101,
  author       = {Khurram Soomro and
                  Amir Roshan Zamir and
                  Mubarak Shah},
  title        = {{UCF101}: A Dataset of 101 Human Actions Classes From Videos in The Wild},
  journal      = {arXiv preprint arXiv:1212.0402},
  year         = {2012},
}

@inproceedings{oxfordpets,
  author       = {Omkar M. Parkhi and
                  Andrea Vedaldi and
                  Andrew Zisserman and
                  C. V. Jawahar},
  title        = {Cats and dogs},
  booktitle    = {Computer Vision and Pattern Recognition},
  pages        = {3498--3505},
  year         = {2012},
}

@inproceedings{flowers102,
  author       = {Maria{-}Elena Nilsback and
                  Andrew Zisserman},
  title        = {Automated Flower Classification over a Large Number of Classes},
  booktitle    = {Indian Conference on Computer Vision, Graphics and Image Processing},
  pages        = {722--729},
  year         = {2008},
}

@inproceedings{dtd,
  author       = {Mircea Cimpoi and
                  Subhransu Maji and
                  Iasonas Kokkinos and
                  Sammy Mohamed and
                  Andrea Vedaldi},
  title        = {Describing Textures in the Wild},
  booktitle    = {Computer Vision and Pattern Recognition},
  pages        = {3606--3613},
  year         = {2014},
}

@inproceedings{sun397,
  author       = {Jianxiong Xiao and
                  James Hays and
                  Krista A. Ehinger and
                  Aude Oliva and
                  Antonio Torralba},
  title        = {SUN database: Large-scale scene recognition from abbey to zoo},
  booktitle    = {Computer Vision and Pattern Recognition},
  pages        = {3485--3492},
  year         = {2010},
}

@inproceedings{stanfordcars,
  author       = {Jonathan Krause and
                  Michael Stark and
                  Jia Deng and
                  Li Fei{-}Fei},
  title        = {3D Object Representations for Fine-Grained Categorization},
  booktitle    = {International Conference on Computer Vision Workshops},
  pages        = {554--561},
  year         = {2013},
}

@article{fgvcaircraft,
  author       = {Subhransu Maji and
                  Esa Rahtu and
                  Juho Kannala and
                  Matthew B. Blaschko and
                  Andrea Vedaldi},
  title        = {Fine-Grained Visual Classification of Aircraft},
  journal      = {arXiv:1306.5151},
  year         = {2013},
}

@inproceedings{food101,
  author       = {Lukas Bossard and
                  Matthieu Guillaumin and
                  Luc Van Gool},
  title        = {{Food-101} - Mining Discriminative Components with Random Forests},
  booktitle    = {European Conference on Computer Vision},
  pages        = {446--461},
  year         = {2014},
}

@article{Schuhmann2021LAION400MOD,
  title        = {{LAION-400M}: Open Dataset of {CLIP}-Filtered 400 Million Image-Text Pairs},
  author       = {Christoph Schuhmann and
                  Richard Vencu and
                  Romain Beaumont and
                  Robert Kaczmarczyk and
                  Clayton Mullis and
                  Aarush Katta and
                  Theo Coombes and
                  Jenia Jitsev and
                  Aran Komatsuzaki},
  journal      = {arXiv:2111.02114},
  year         = {2021},
}

\appendix
\section{Detailed Derivations}
\label{app:derivations}

\subsection{Reconstruction after Centering and Normalization}
\label{app:reconstruction-derivation}

The preprocessing in \eqnref{eq:preprocessing} separates the centered prompt embedding into
a direction and a scale. Define
\begin{equation}
  \rho_{\bm{z}}=\lVert\bm{z}-\bm{\mu}\rVert_2.
\end{equation}
Provided that $\rho_{\bm{z}}>0$, multiplying \eqnref{eq:preprocessing} by $\rho_{\bm{z}}$
gives
\begin{equation}
  \bm{z}-\bm{\mu}
  =\rho_{\bm{z}}\tilde{\bm{z}},
  \quad
  \bm{z}=\bm{\mu}+\rho_{\bm{z}}\tilde{\bm{z}}.
  \label{eq:inverse-preprocessing-app}
\end{equation}
Let the decomposition residual be
\begin{equation}
  \bm{e}=\tilde{\bm{C}}\bm{w}^{*}-\tilde{\bm{z}}.
\end{equation}
If the discarded scale is retained, the unnormalized reconstruction is
\begin{equation}
  \begin{aligned}
  \widehat{\bm{u}}(\rho_{\bm{z}})
  =\bm{\mu}+\rho_{\bm{z}}
    \tilde{\bm{C}}\bm{w}^{*}
  =\bm{z}+\rho_{\bm{z}}\bm{e}.
  \end{aligned}
  \label{eq:scale-aware-raw-reconstruction}
\end{equation}
Hence, before the final unit normalization,
\begin{equation}
  \lVert\widehat{\bm{u}}(\rho_{\bm{z}})-\bm{z}\rVert_2
  =\rho_{\bm{z}}\lVert\bm{e}\rVert_2.
  \label{eq:reconstruction-residual}
\end{equation}
The corresponding scale-aware unit vector is
\begin{equation}
  \widehat{\bm{z}}(\rho_{\bm{z}})
  =\frac{
    \bm{\mu}+\rho_{\bm{z}}\tilde{\bm{C}}\bm{w}^{*}
  }{
    \lVert\bm{\mu}+\rho_{\bm{z}}
    \tilde{\bm{C}}\bm{w}^{*}\rVert_2
  }.
  \label{eq:scale-aware-reconstruction}
\end{equation}
When $\bm{e}=\bm{0}$ and $\bm{z}$ is unit-normalized,
\eqnref{eq:scale-aware-reconstruction} recovers $\bm{z}$ exactly. The direction-only
reconstruction in \eqnref{eq:reconstruction} is obtained by setting $\rho_{\bm{z}}=1$. It
therefore reconstructs the direction used by the decomposition but is not, in general, an
exact inverse of the preprocessing operation.

\subsection{Local Sensitivity in Concept Coordinates}
\label{app:gradient-derivation}

For one training example, let
\begin{equation}
  p_k=\frac{\exp(s_k/\tau)}
  {\sum_{\ell}\exp(s_{\ell}/\tau)},
  \quad
  \mathcal{L}=-\sum_k t_k\log p_k,
\end{equation}
where $\sum_k t_k=1$. Substituting the softmax into the loss gives
\begin{equation}
  \mathcal{L}
  =-\frac{1}{\tau}\sum_k t_k s_k
  +\log\sum_{\ell}\exp(s_{\ell}/\tau).
\end{equation}
Differentiating each term with respect to $s_i$ yields
\begin{equation}
  \begin{aligned}
  \frac{\partial\mathcal{L}}{\partial s_i}
  =-\frac{t_i}{\tau}
  +\frac{1}{\tau}
  \frac{\exp(s_i/\tau)}
  {\sum_{\ell}\exp(s_{\ell}/\tau)}
  =\frac{p_i-t_i}{\tau}.
  \end{aligned}
  \label{eq:score-gradient-app}
\end{equation}

Next, set $r_i=\lVert\bm{u}_i\rVert_2$ and $\bm{z}_i=\bm{u}_i/r_i$. The differential of the
norm is
\begin{equation}
  \mathrm{d}r_i
  =\frac{\bm{u}_i^{\top}\mathrm{d}\bm{u}_i}{r_i}
  =\bm{z}_i^{\top}\mathrm{d}\bm{u}_i.
  \label{eq:norm-differential}
\end{equation}
Using the product rule on $\bm{z}_i=\bm{u}_i r_i^{-1}$ gives
\begin{equation}
  \begin{aligned}
  \mathrm{d}\bm{z}_i
  =\frac{\mathrm{d}\bm{u}_i}{r_i}
  -\frac{\bm{u}_i}{r_i^2}\,\mathrm{d}r_i
  =\frac{1}{r_i}
  \left(\bm{I}-\bm{z}_i\bm{z}_i^{\top}\right)
  \mathrm{d}\bm{u}_i.
  \end{aligned}
  \label{eq:normalization-differential}
\end{equation}
Because $\bm{u}_i=\tilde{\bm{C}}\bm{w}_i+\bm{\mu}$,
\begin{equation}
  \frac{\partial\bm{u}_i}{\partial w_{ij}}
  =\tilde{\bm{c}}_j.
  \label{eq:u-gradient-app}
\end{equation}
It follows from $s_i=\bm{f}^{\top}\bm{z}_i$ that
\begin{equation}
  \frac{\partial s_i}{\partial w_{ij}}
  =\frac{1}{\lVert\bm{u}_i\rVert_2}
  \bm{f}^{\top}
  \left(\bm{I}-\bm{z}_i\bm{z}_i^{\top}\right)
  \tilde{\bm{c}}_j.
  \label{eq:similarity-gradient-app}
\end{equation}
Multiplying \eqnref{eq:score-gradient-app} and \eqnref{eq:similarity-gradient-app} gives
\eqnref{eq:activation-gradient}. Equivalently, the gradient with respect to the entire
concept-coordinate vector is
\begin{equation}
  \nabla_{\bm{w}_i}\mathcal{L}
  =\frac{p_i-t_i}
  {\tau\lVert\bm{u}_i\rVert_2}
  \tilde{\bm{C}}^{\top}
  \left(\bm{I}-\bm{z}_i\bm{z}_i^{\top}\right)
  \bm{f}.
  \label{eq:activation-vector-gradient}
\end{equation}

\begin{figure*}[!t]
  \centering
  \begin{minipage}[t]{0.485\textwidth}
    \centering
    \includegraphics[width=\linewidth]{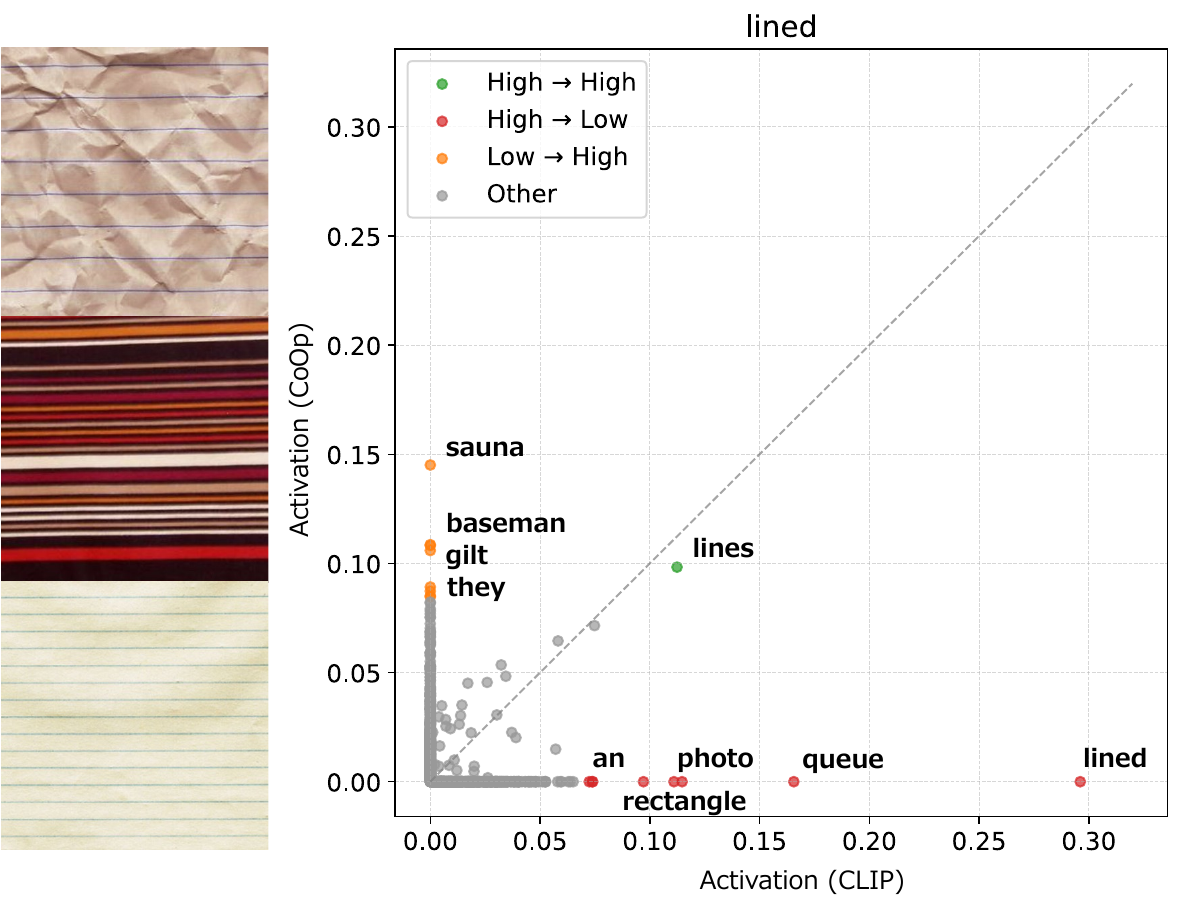}
    \small (a) DTD: ``lined''
  \end{minipage}\hfill
  \begin{minipage}[t]{0.485\textwidth}
    \centering
    \includegraphics[width=\linewidth]{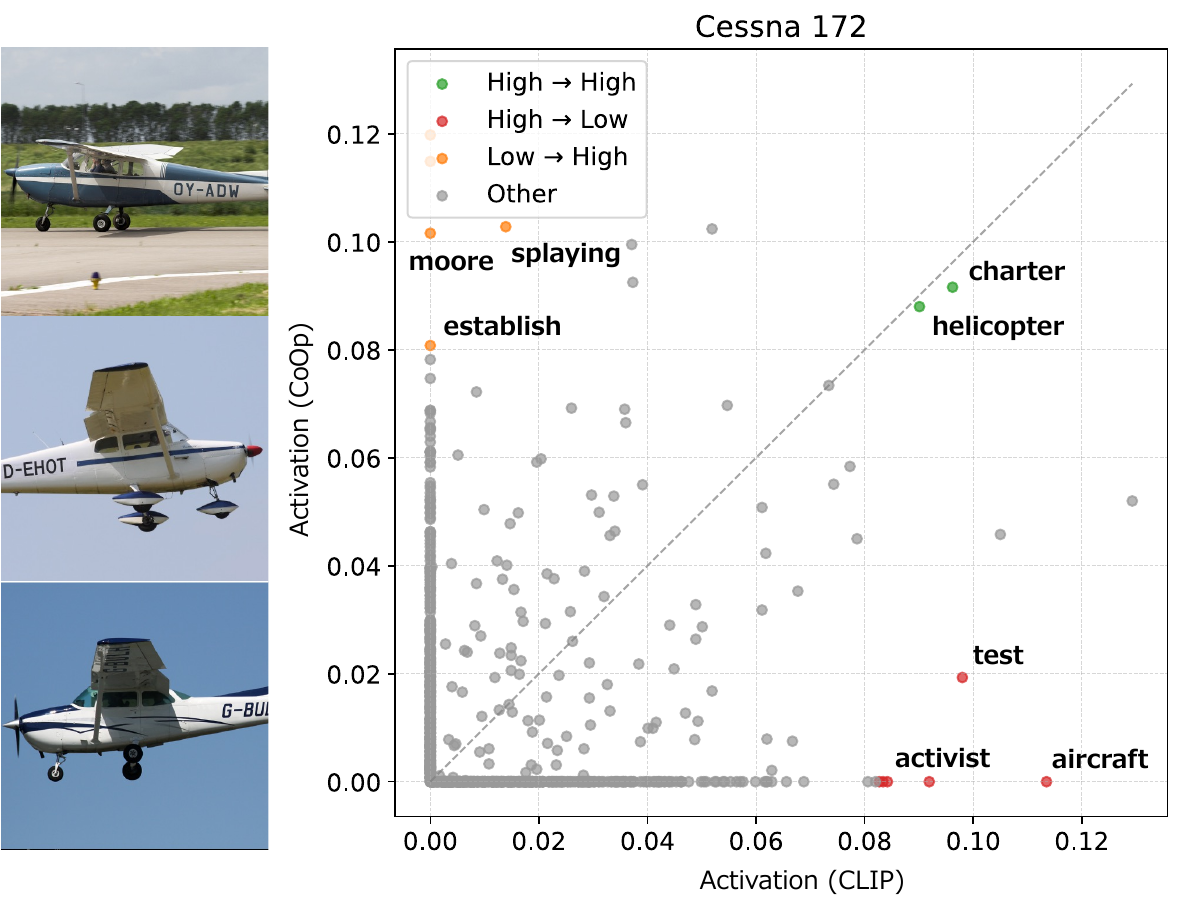}
    \small (b) FGVCAircraft: ``Cessna 172''
  \end{minipage}

  \vspace{0.8em}
  \begin{minipage}[t]{0.485\textwidth}
    \centering
    \includegraphics[width=\linewidth]{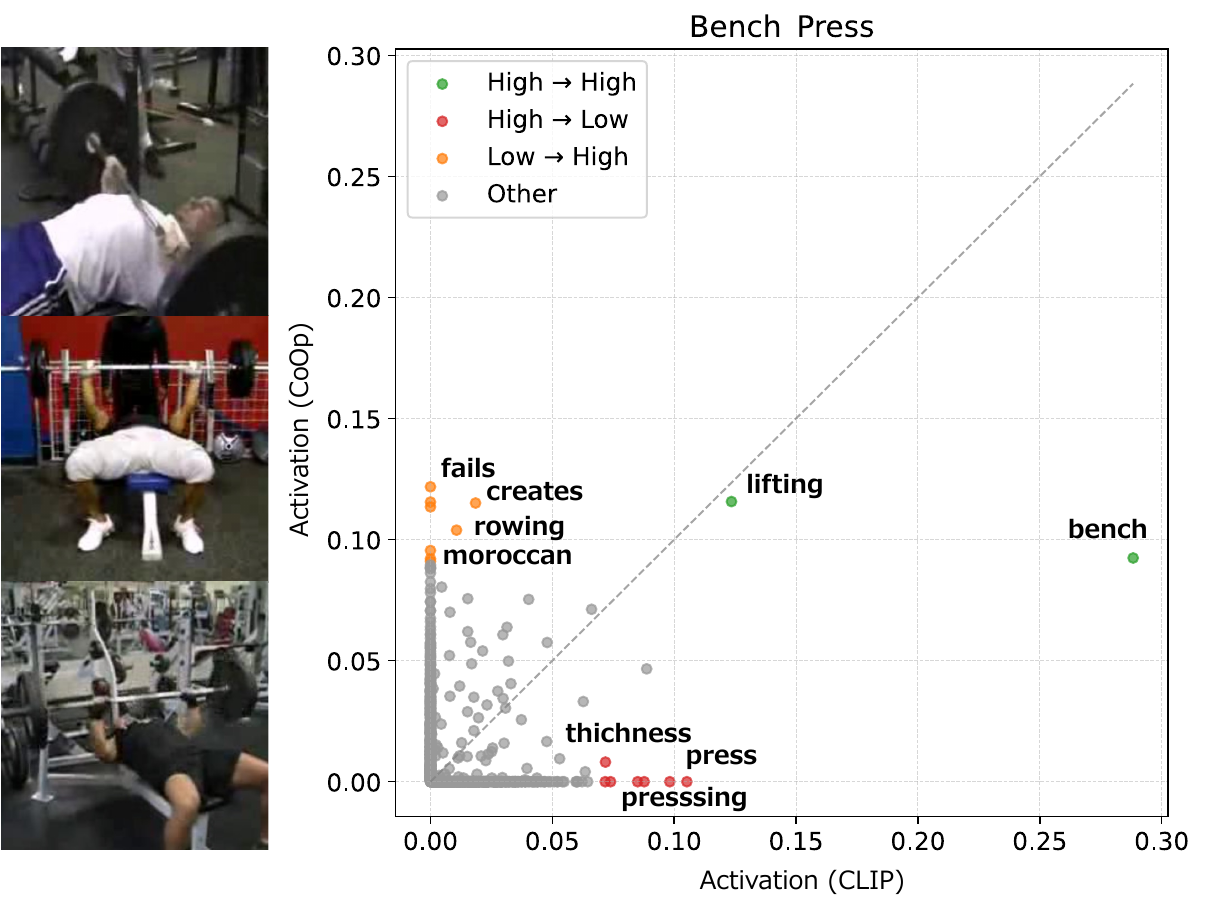}
    \small (c) UCF101: ``Bench Press''
  \end{minipage}\hfill
  \begin{minipage}[t]{0.485\textwidth}
    \centering
    \includegraphics[width=\linewidth]{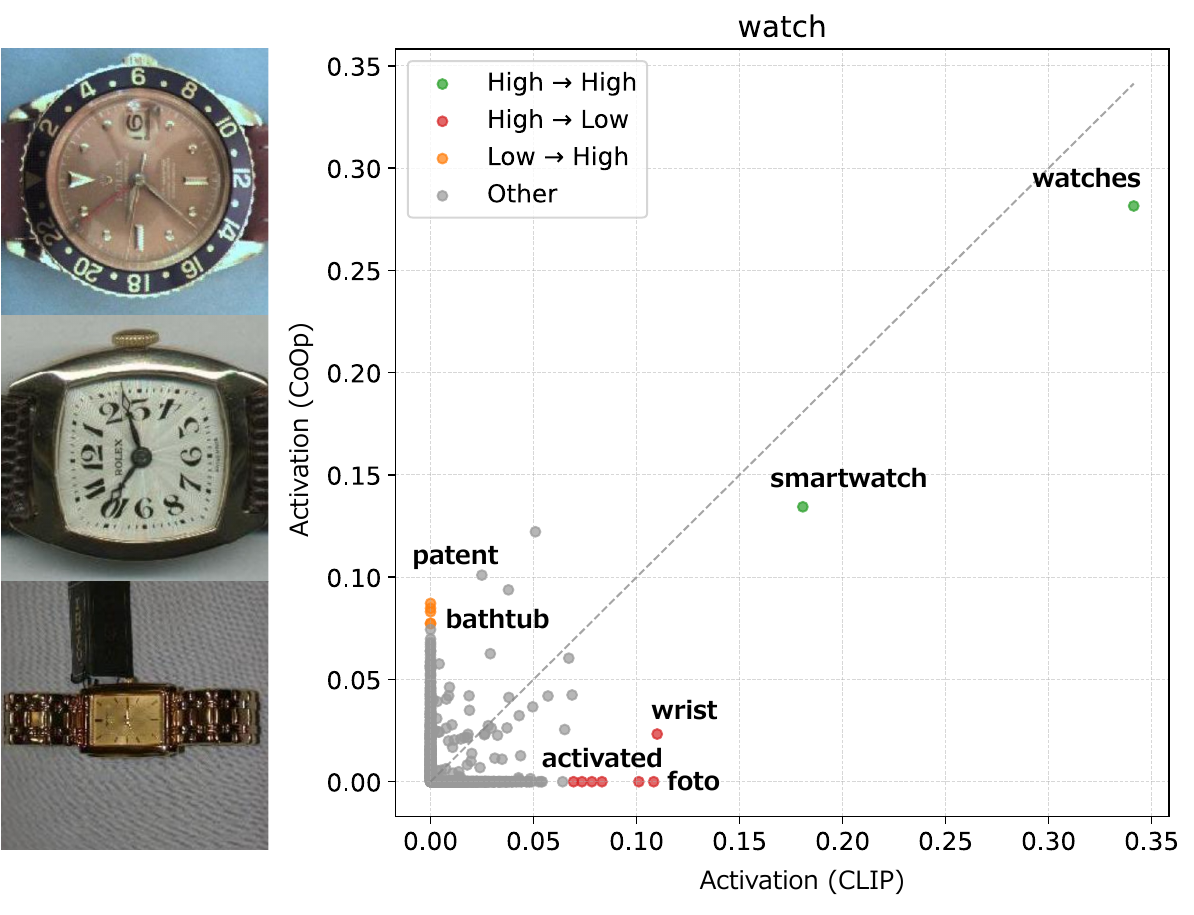}
    \small (d) Caltech101: ``watch''
  \end{minipage}
  \caption{Additional decomposition-coefficient changes for DTD,
  FGVCAircraft, UCF101, and Caltech101. Colors denote
  High$\rightarrow$High, High$\rightarrow$Low, and
  Low$\rightarrow$High transitions.}
  \label{fig:supp-cases-1}
\end{figure*}

This calculation treats $\bm{w}_i$ as a local coordinate vector and assumes that it affects
only $s_i$. 
The underlying prompt-learning method instead optimizes prompt parameters $\bm{\theta}$,
which may be shared across classes.
If $\bm{J}_i=\partial\bm{z}_i/\partial\bm{\theta}$ denotes the text-embedding Jacobian, its
parameter gradient has the more general form
\begin{equation}
  \nabla_{\bm{\theta}}\mathcal{L}
  =\sum_i\bm{J}_i^{\top}
  \nabla_{\bm{z}_i}\mathcal{L},
  \label{eq:coop-parameter-gradient}
\end{equation}
where $\bm{J}_i$ includes the text-encoder Jacobian and any cross-class parameter sharing.
Therefore, \eqnref{eq:activation-gradient} should be interpreted as a class-wise,
embedding-level sensitivity expressed in the PromptSpLiCE coordinate system. It is not the
total derivative of the post-hoc solution $\bm{w}_i^{*}(\bm{z}_i(\bm{\theta}))$, which
would additionally require differentiating through the Lasso solution and accounting for
changes in active support. At a nonnegativity boundary $w_{ij}=0$, it describes the local
unconstrained derivative; feasible perturbations of the coefficient remain one-sided.

\begin{figure*}[!t]
  \centering
  \begin{minipage}[t]{0.485\textwidth}
    \centering
    \includegraphics[width=\linewidth]{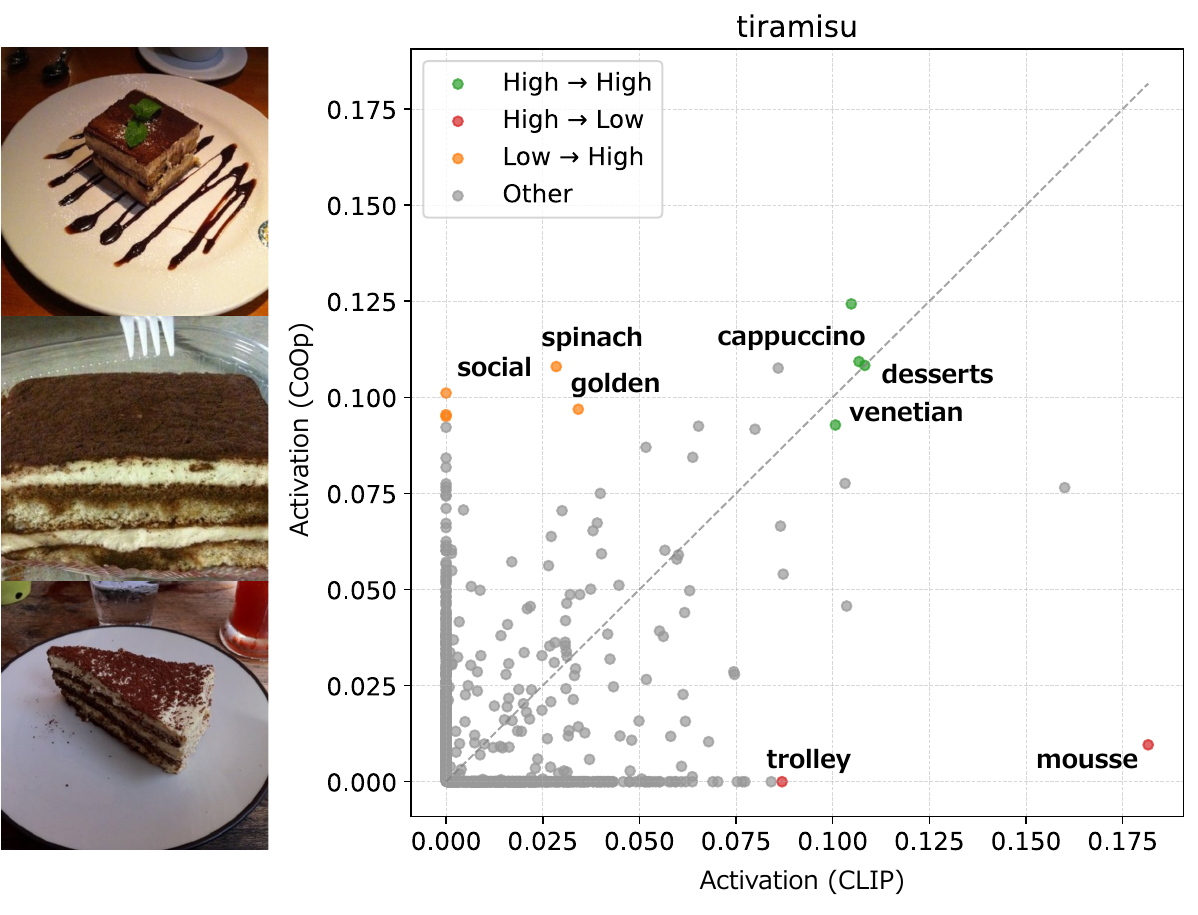}
    \small (a) Food101: ``tiramisu''
  \end{minipage}\hfill
  \begin{minipage}[t]{0.485\textwidth}
    \centering
    \includegraphics[width=\linewidth]{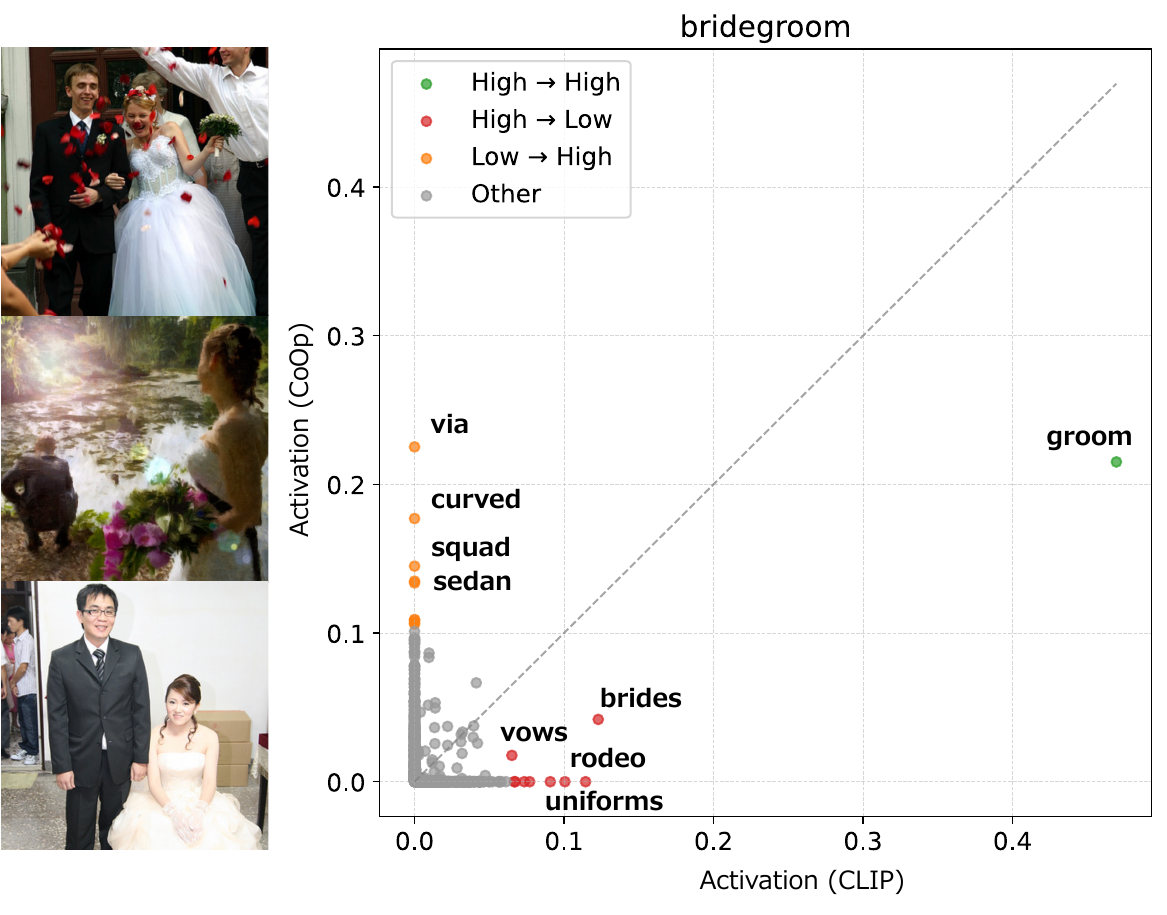}
    \small (b) ImageNet: ``bridegroom''
  \end{minipage}
  \caption{Additional decomposition-coefficient changes for Food101 and
  ImageNet.}
  \label{fig:supp-cases-2}
\end{figure*}

\begin{figure}[t]
  \centering
  \includegraphics[width=0.5\textwidth]
    {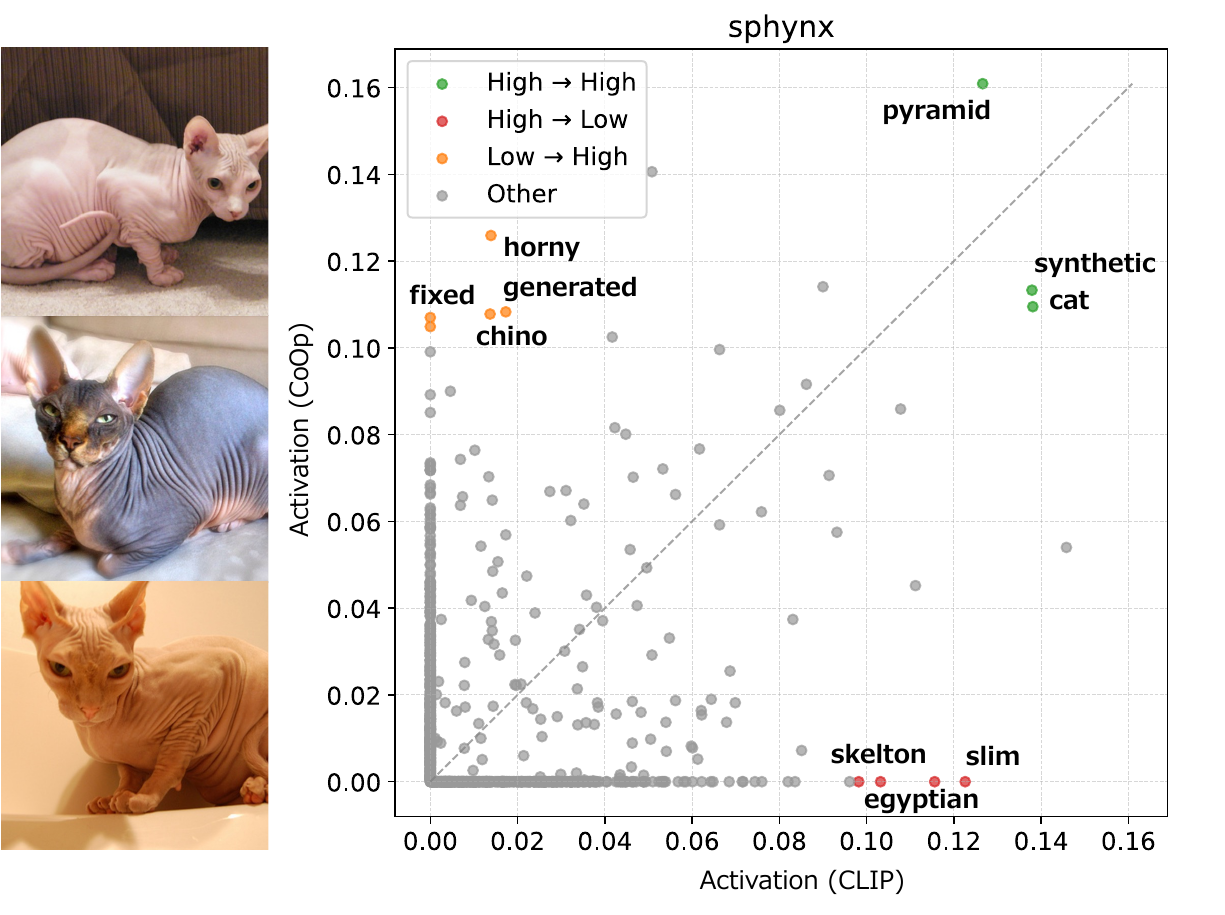}
  \caption{Additional decomposition-coefficient changes for the OxfordPets
  class ``sphynx.''}
  \label{fig:supp-cases-3}
\end{figure}

\section{Additional Qualitative Results}
\label{app:additional-cases}

Figures~\ref{fig:supp-cases-1}--\ref{fig:supp-cases-3} show the
remaining seven datasets not included in \figref{fig:case-study} or
\figref{fig:main-additional-cases}, giving qualitative coverage of
all 11 datasets. Each panel pairs representative images with a
scatter plot whose horizontal and vertical axes show fitted
coefficients before (CLIP) and after (CoOp) learning, respectively.
Green, red, and orange points denote High$\rightarrow$High,
High$\rightarrow$Low, and Low$\rightarrow$High transitions; gray
points show the remaining dictionary terms. These transitions vary across
datasets, consistent with \tabref{tab:rank-transitions}. The
discussion below focuses on the labeled terms selected to
illustrate the three transition types; these labels are examples
rather than an exhaustive or unique summary of each coefficient distribution.

\paragraph{DTD and FGVCAircraft.}
For DTD's ``lined'' class, \emph{lines} remains highly ranked, whereas
\emph{lined}, \emph{photo}, \emph{rectangle}, and \emph{queue} move
downward. Terms such as \emph{sauna}, \emph{baseman}, and \emph{gilt}
move upward. For
FGVCAircraft's ``Cessna 172'' class, \emph{charter} and
\emph{helicopter} remain highly ranked, while \emph{aircraft} moves
downward and \emph{moore}, \emph{splaying}, and \emph{establish} move
upward. Thus, category-related and less intuitive dictionary terms
can undergo different rank transitions.

\paragraph{UCF101 and Caltech101.}
For UCF101's ``Bench Press'' class, \emph{lifting} and \emph{bench}
remain highly ranked, but \emph{press} and \emph{pressing} move
downward, while \emph{fails}, \emph{creates}, \emph{rowing}, and
\emph{moroccan} move upward. For Caltech101's ``watch'' class,
\emph{watches} and \emph{smartwatch} remain highly ranked, whereas
\emph{wrist} moves downward and \emph{patent} and \emph{bathtub} move
upward. These cases show that substantial coefficient-rank turnover
can coexist with the retention of highly diagnostic terms.

\paragraph{Food101 and ImageNet.}
For Food101's ``tiramisu'' class, \emph{desserts} and
\emph{cappuccino} remain highly ranked, while \emph{mousse} moves
downward and \emph{social}, \emph{spinach}, and \emph{golden} move
upward. For ImageNet's ``bridegroom'' class, \emph{groom} remains
dominant, whereas \emph{brides}, \emph{vows}, and \emph{uniforms} move
downward and \emph{via}, \emph{curved}, \emph{squad}, and \emph{sedan}
move upward. These cases again show that a central category-related
term can remain highly ranked while related and less intuitive terms
undergo different transitions.

\paragraph{OxfordPets.}
For OxfordPets' ``sphynx'' class, \emph{cat}, \emph{synthetic}, and
\emph{pyramid} remain highly ranked. In contrast, \emph{egyptian} and
\emph{slim} move downward, while \emph{horny}, \emph{fixed},
\emph{generated}, and \emph{chino} move upward. Thus, broad and
fine-grained labels can behave differently in the fitted coefficient
ranking.

Across these seven cases, at least one intuitive term often remains
in the High$\rightarrow$High group. At the same time, related terms
can move to High$\rightarrow$Low, while Low$\rightarrow$High
frequently contains labels with no clear human relation to the class.
This pattern is consistent with \tabref{tab:rank-transitions} and
illustrates why high full-embedding reconstruction fidelity does not,
by itself, establish stability or semantic fidelity of the displayed
top-ranked terms.
\end{document}